# Securing generative artificial intelligence with parallel magnetic tunnel junction true randomness

*Youwei Bao, Shuhan Yang, and Hyunsoo Yang\**

Department of Electrical and Computer Engineering, National University of Singapore,
Singapore 117583, Singapore
E-mail: eleyang@nus.edu.sg

Deterministic pseudo random number generators (PRNGs) used in generative artificial intelligence (GAI) models produce predictable patterns vulnerable to exploitation by attackers. Conventional defenses against the vulnerabilities often come with significant energy and latency overhead. Here, we embed hardware-generated true random bits from spin-transfer torque magnetic tunnel junctions (STT-MTJs) to address the challenges. A highly parallel, FPGA-assisted prototype computing system delivers megabit-per-second true random numbers, passing NIST randomness tests after in-situ operations with minimal overhead. Integrating the hardware random bits into a generative adversarial network (GAN) trained on CIFAR-10 reduces insecure outputs by up to 18.6 times compared to the low-quality random number generators (RNG) baseline. With nanosecond switching speed, high energy efficiency, and established scalability, our STT-MTJ-based system holds the potential to scale beyond $10^6$ parallel cells, achieving gigabit-per-second throughput suitable for large language model sampling. This advancement highlights spintronic RNGs as practical security components for next-generation GAI systems.

## 1. Introduction

Generative artificial intelligence (GAI), which uses machine learning models to generate text,[1] images,[2] music,[3] videos,[4] and complex simulations,[5] has transformed industries and reshaped daily life. It has achieved remarkable commercial success across various applications, including ChatGPT, Midjourney, and Sora. However, the rapid advancements outpace the development of security measures. Previous studies[6] documented numerous attacks, and more are expected in real-world scenarios. Research indicates that attackers can bypass built-in security measures in models like GPT-4 without direct access to the internal model structure. Simply by interacting with the model application programming interface (API) and fine-tuning it with as few as 15 harmful examples, attackers can make the model generate malicious content.[7] A common aspect of these exploits is the deterministic nature of pseudo random number generators (PRNGs),[8] which supply latent codes, diffusion noise, or token sampling. Since the PRNG produces identical sequences for a given seed, the attacker can iteratively query the model, analyze the relationship between inputs and outputs, estimate the underlying gradients, and gradually direct the model to generate harmful content.[6a] Adversarial training[9] is a conventional defense strategy for traditional machine learning models. It expands datasets with adversarial examples and optimizes the entire network. By learning to handle challenging adversarial examples correctly, models become more robust against future attacks. However, generating adversarial examples and training the models using them leads to overall durations 3 to 30 times longer,[10] which substantially increases energy consumption and costs, making the strategy impractical for large-scale GAI models.

Instead of training models to handle harmful examples, a more cost-effective strategy is to reduce vulnerability during inference.[11] As illustrated in Figure 1a, during black-box attacks, attackers systematically probe the model by varying user inputs, such as seed values, and observing the outputs. This process enables attackers to understand how user inputs correspond to specific outputs, as consistent patterns emerge. For example, certain seeds consistently yield specific features, providing attackers insights into confidential training data or enabling them to generate specific harmful outputs. The deterministic nature creates a fixed dependency chain from user input (e.g., seed) through the PRNG-generated random sequence (latent code) to the model output (e.g., generated image). Moreover, if the PRNG demonstrates biases that narrow the output range, attackers can focus on a narrower subset, amplifying their attack success rate.[12] In contrast, the inherent unpredictability of true random number generators (TRNGs) disrupts the fixed dependency chain, preventing attackers from consistently crafting inputs that yield predictable patterns.

Among TRNGs,[13] spin-transfer torque magnetic tunnel junctions (STT-MTJs) stand out due to their nanosecond switching speeds enabling mega to giga bit per second throughput,[14] sub-picojoule energy per bit,[15] and mature fabrication processes with demonstrated giga-size array densities.[16] Despite promising progress highlighting these MTJ advantages, research remains largely device-centric,[17] without connecting to system-level AI performance.[18] To bridge the gap, we implement a 16-MTJ system and interface with a field-programmable gate array (FPGA). Our system architecture enables parallel individual access to all MTJ cells. Therefore, the bit-per-second, which serves as the key performance metric, scales linearly with the number of MTJ cells that can be accessed in parallel. Due to fluctuations of environmental and measurement conditions, the raw MTJ outputs exhibit reduced randomness quality. Thus, we implement two post-processing schemes: a lightweight bitwise XOR between raw bits,[19] and a standard Toeplitz hash meeting cryptographic randomness standard.[17d, 20] Both processed streams pass NIST SP 800-22 tests[8] with 100% success. In addition, we embed the TRNG into the latent-code generator of a deep generative adversarial network (GAN) trained on CIFAR-10. Inference using enhanced MTJ-based latent codes reduces low-diversity (biased insecure) image clusters by up to 18.6× across 10,000 generated images per RNG. Although our experimental focus is a GAN, the same entropy source naturally applies to other GAI workflows. On edge devices, the high throughput and low-power MTJ entropy facilitates on-device generation without cloud exposure, aligning with AI safety frameworks.[21] Our work establishes STT-MTJ TRNGs as practical, high performance, and energy-efficient building blocks for secure and responsible GAI.

## 2. Results and Discussion
### 2.1. Constructing the MTJ-based computing system

An STT-MTJ consists of two ferromagnetic layers separated by a thin insulating barrier. One layer has a fixed magnetization, while the other, known as the free layer, can switch its magnetic orientation. The relative alignment of the ferromagnetic layers determines the resistance of the MTJ, resulting in two stable states: parallel (P), with a low resistance, and antiparallel (AP), with a high resistance, representing binary values 1 and 0. STT is used to switch the MTJ between the two states by applying a current that manipulates the magnetic moment of the free layer. In conventional memory applications, controlled current pulses are used to deterministically set the MTJ into a specific state. However, by carefully tuning the amplitude and duration of the current pulses, the MTJ can be driven into a metastable regime where thermal fluctuations determine the final state.[13b, 17a] As shown in Figure 1b, we tune the switching probability by varying the amplitude of voltage pulses $V_{dd}$ (the pulse width is fixed at 5 μs). The symbols depict a sigmoid-like switching probability response to different input amplitudes from 16 individual MTJs. The switching probability is calculated by averaging the MTJ outputs over 1,000 input pulses. The stochastic switching behavior enables the MTJ to function as a TRNG, as the output is inherently unpredictable and governed by thermal noise.

Based on this mechanism, we construct a system that consists of 16 MTJs with a peripheral circuit and an FPGA, as illustrated in Figure 1c. A 16-channel digital-to-analog converter (DAC) supplies 16 analog input voltages to control the switching probability of each individual MTJ, while a 16-channel analog-to-digital converter (ADC) samples the resulting state of each MTJ (binarized using a predefined threshold) after each input pulse. The setup enables the parallel usage of all 16 MTJ cells, with the output raw bits transmit to and store in the FPGA. For our generation task, we set the switching probability of all computing cells to ~50%, as shown in Figure 1d. Further details on the development and characterization of our computing system are provided in the Experimental Section. In practice, the ADC used in our prototype could be replaced by read sense amplifiers (RSAs),[22] which are well established in magnetic random-access memory technology and offer significantly lower energy costs and smaller footprints. Thus, by utilizing the intrinsic stochastic switching behavior of MTJ devices, we can generate random numbers, which can be further optimized for resource-efficient operation.

2.2. **Statistical property analysis of RNGs**

After obtaining the raw random bit sequence, we verify the functionality of our MTJ as an RNG. Each 32-bit binary output is converted to its decimal equivalent, and a histogram based on $10^7$ bits is shown in Figure 2a. Despite fine tuning, the raw data are clearly not uniformly distributed. As the thermal noise such as device heating is unlikely to remain constant throughout the experiment, and even slight variations can skew the distribution of output bits away from uniformity. Moreover, assuming complete independence between raw output bits would be unrealistic. Imperfections in the experimental setup, such as noisy read circuits, may introduce correlations, further degrading the quality of randomness. To address these issues, we apply two established post-processing techniques (Section S1, Supporting Information). The first is a lightweight scheme with negligible energy overhead, where we use intrinsic memory operations to perform a bitwise XOR across three raw bits (MTJ with XOR). The second is a standard extraction scheme using Toeplitz hashing (MTJ extraction), a member of the universal hashing family that offers provably secure randomness extraction. To reduce the computational cost of Toeplitz hashing, we adopt the fast Fourier transform (FFT). The 32-bit binary outputs from the post-processed random sequence are again converted to decimal form, and the resulting histograms are shown in Figure 2b,c. These distributions exhibit visually satisfactory uniformity. For comparison, we also implement a linear feedback shift register (LFSR), a commonly used resource-efficient PRNG. Its output histogram, shown in Figure 2d, demonstrates good uniformity.

Next, we evaluate all four RNGs using the NIST SP 800-22 Statistical Test Suite (NIST tests),[8] a standard consisting of 15 tests designed to assess the statistical quality of random bitstreams. Following the recommended sequence length, we test $5.5\times10^7$ bits of random data for each RNG. As summarized in Table 1, both MTJ-based TRNG schemes (MTJ with XOR and MTJ extraction) pass all tests, demonstrating high-quality randomness and strong security properties. Moreover, the uniformity-of-p-values of the two schemes cover the full [0, 1] range: from 0.0002 to 0.9943 for MTJ with XOR, and from 0.0006 to 0.9879 for MTJ extraction. Over 96% of the values fall within the [0.1, 0.9] range, with only 7 and 4 sub-tests below 0.01, respectively, and none below 0.0001. This indicates a uniform distribution, reflecting robust randomness rather than a small number of lucky passes. In contrast, the MTJ raw data and the LFSR fail to pass all the tests. As discussed earlier, the failure of the raw MTJ output is expected due to its non-uniform distribution and bit correlations. Although the LFSR output appears visually uniform, it has fundamental limitations.[23] LFSRs generate sequences through linear operations on previous bits, resulting in low linear complexity. This causes failures in the linear complexity test. Moreover, structured dependencies among bits can reduce the rank of disjoint sub-matrices constructed during the rank test, leading to failure in this test as well. These limitations degrade randomness quality and thus compromise security.

Notably, advanced general-purpose PRNGs, such as Xoroshiro128+, may pass all NIST tests. However, their deterministic nature opens the door to black-box attacks at the application level. As a secure alternative, cryptographically secure PRNGs (CSPRNGs) are designed to generate highly unpredictable numbers. However, a CSPRNG is significantly more complex and resource-intensive than general-purpose PRNGs. It requires additional components such as ciphers, cryptographic hash functions, runtime tests, and robust seeding mechanisms that rely on a true random seed to initiate a secure sequence.[24] In contrast, our enhanced MTJ-based TRNGs not only demonstrate strong statistical properties by passing all NIST tests but also produce genuinely unpredictable outputs. In particular, our lightweight scheme has negligible energy overhead while significantly improving randomness, which makes it a promising and resource-efficient solution for AI security applications.

### 2.3. Image generation with MTJ-based TRNGs

We further implement a GAN that integrates our hardware RNGs. Compared to their early predecessors,[25] recent GANs have significantly advanced in their ability to generate realistic and diverse images, benefiting from improvements in architectural design and training strategies.[26] However, the fundamental structure of GANs remains largely unchanged, consisting of two main components: a generator and a discriminator, as illustrated in Figure 3a. During image generation, GANs use a vector of random numbers (latent code) as input to seed stochastic image synthesis. The generator progressively transforms the latent code into human-interpretable images through multiple intermediate layers. Apart from the latent code, all other operations within the model are deterministic. Figure 3a shows an example of a dog image generated from the initial latent code, through intermediate-layer representations, to the final output. By varying the latent code, the GAN produces diverse images. The discriminator performs two tasks: class prediction, to identify the class label of the input image using ground truth labels from datasets like CIFAR-10,[27] and adversarial prediction, to determine whether the image is real (from dataset) or generated by the model. During training, the discriminator predictions are iteratively compared against their respective targets, and backpropagation is used to update the weights of both the generator and the discriminator (Figure 3a). Specifically, we implement an autoencoder-based discriminator and adopt a training strategy designed to enhance image quality (Table S1 and Section S2, Supporting Information). The model is pre-trained on a GPU and demonstrates superior performance compared to classical GANs.[26a, 28]

Next, we replace the algorithmic RNG with our hardware-based RNGs during inference for image generation. The latent code for each generated image is a vector of length 110, comprising 100 random numbers in the range [-1, 1], along with an additional 10 numbers encoding class information. To generate each random number in the latent code vector, we take a full 32-bit word from our random bit sequence, producing a binarized sequence. First, we convert the binary value to decimal, and then we normalize it to the [-1, 1] range. This normalized value is then fed into the GAN via a Python interface in single-precision floating-point format. Producing 100 random numbers consumes a total of 3,200 bits to generate a single image. In Figure 3b, we illustrate a latent code matrix for generating a batch of 100 images. Subsequently, we generate 10,000 images (each 64×64 pixels) using all our hardware-based RNGs.

To evaluate model performance, we calculate the Inception score[29] (IS) for the generated images, a widely adopted metric for assessing generative models. The IS evaluates both the objectness (i.e., each image is classified with high certainty) and the class diversity (i.e., the classifier used for evaluation identifies a wide variety of classes among images) of generated images. The results, shown in Figure 3c, demonstrate that our model achieves an IS as high as 10.28 (the higher, the better). This score closely approaches the performance of recent state-of-the-art models[30] and is not far below the IS of the CIFAR-10 training dataset itself (11.24), which typically represents an upper boundary for models trained on the same dataset. Among all tested hardware RNGs, we observe similar levels of IS. However, evaluating generative models quantitatively remains challenging. Although the IS is a popular metric, it may overlook certain aspects.[26a] Therefore, additional metrics, such as learned perceptual image patch similarity (LPIPS), are analyzed in the following section. Figure 3d provides examples of images generated using our MTJ with XOR TRNG, each clearly matching its corresponding class. Details of the IS evaluation and additional image results for each RNG are in Figure S1 and Section S3, Supporting Information.

## 2.4. System-level analysis of RNGs

By carefully observing the generated images, we notice repetitive patterns in the results produced by low-quality RNGs (MTJ raw data and LFSR). Examples of visually similar image pairs are presented in Figure 4a. Intuitively, we perceive similarities within each pair, particularly regarding color and the shape of main objects (such as a deer or a car in the examples provided). However, these pairs are subtly different, indicating correlation rather than exact duplication. To better understand the observed correlations, we extract feature maps from similar image pairs (see details in Section S4, Supporting Information). Figure 4a shows low-level feature maps, which capture basic image attributes that are intuitively interpretable by humans, while Figure 4b displays high-level feature maps, capturing more abstract patterns meaningful to AI systems. As expected, the low-level features within each image pair closely resemble each other, aligning with human perception. Interestingly, high-level features also exhibit similar patterns, with comparable values appearing at similar spatial locations. Our observations indicate the presence of correlations within the latent space, which are directly influenced by the quality of the underlying RNG. Given that the IS primarily measures class diversity rather than intra-class diversity,[31] we further investigate diversity among images generated within the same class. This is important since biased model outputs can directly compromise security.

To systematically evaluate image diversity, we apply t-distributed stochastic neighbour embedding[32] (t-SNE), a statistical method for visualizing high-dimensional data (i.e., 64×64 pixel images) by mapping them into a lower-dimensional (typically two-dimensional) space. t-SNE preserves local relationships between the data; points close in the original high-dimensional space remain nearby after dimensionality reduction, while distant points remain separated. Figure 4c displays t-SNE plots from images generated by a low-quality RNG (LFSR, left panel) and a high-quality RNG (MTJ with XOR, right panel), where each generated image corresponds to a single data point (10,000 points per plot). Smaller distances between points indicate greater similarity. Images generated by the LFSR form tightly clustered patterns, indicating lower diversity, whereas those from the MTJ with XOR RNG show more scattered and loosely structured distributions, suggesting higher intra-class diversity. Details of the t-SNE are included in Section S5 and Figure S2 of Supporting Information.

Furthermore, we use LPIPS[33] to quantitatively measure similarity between image pairs. LPIPS scores range from 0 to 1, with lower scores indicating greater similarity; we consider image pairs with scores below 0.3 to be perceptually similar (and thus correlated and lower-quality). We randomly divide 1,000 generated images from the same class into two groups, compute pairwise LPIPS scores, and count the number of similar image pairs (Sections S6 and S7, Supporting Information; Figure S3 to S5). As illustrated in Figure 4d, our MTJ with XOR RNG achieves performance comparable to the established cybersecurity-grade RNG[17d] (MTJ extraction). A clear performance gap emerges between low-quality and high-quality RNGs, notably between MTJ raw data and its improved counterparts. Specifically, for class 6, the improved MTJ RNGs achieve an 18.6 times improvement over the MTJ raw data RNG. Performance differences vary across image classes; for instance, classes 3 and 5 exhibit small differences between the LFSR and our improved MTJ RNGs. We interpret the observed latent-space correlations as more complex than the statistical correlations typically detected by standard NIST tests, such as the linear complexity test or rank test. Previous studies[34] suggest that the latent codes in GANs behave analogously to biological genes, where each dimension controls specific characteristics of the generated images. These "genes" interact in a complex, non-uniform, and entangled manner. Although an LFSR provides good bit-level uniformity, its inherently linear structure is insufficient for eliminating the deeper, more subtle correlations within the latent space. Consequently, at the application level, an LFSR produces more biased image outputs compared to high-quality TRNGs. These biases are also task-specific and model-dependent, resulting in variability across different image classes. Given that even minor biases can significantly impact security, it is necessary to integrate GAI with high-quality RNGs.

## 3. Conclusion

Compared to general-purpose PRNGs, our improved MTJ TRNGs offer security advantages. General-purpose PRNGs are inherently deterministic and thus, even those with good statistical properties remain vulnerable to attacks. Moreover, achieving good statistical properties typically requires careful design and implementation. As demonstrated in Figure S6 and Section S7, Supporting Information, intentionally poorly designed LFSRs clearly show how easily statistical properties can be compromised. On the other hand, CSPRNGs offer secure solutions but introduce significant energy overhead. Typical CSPRNG implementations consume tens to hundreds of nanojoules per random bit,[24] which is over $10^5$ times more than the ~1 pJ per random bit[15] estimated for our lightweight scheme when scaled to systems with $10^6$ devices (Figure S7 and S8, Table S2 to S4, and Section S8, Supporting Information). The energy efficiency is especially important for constrained IoT devices powered by small batteries, where energy depletion attacks may even occur.[24]

In summary, we experimentally implement a hardware RNG system based on MTJs with CMOS for random sequence generation. We explore two enhancement schemes to improve statistical properties: a lightweight scheme using in-situ operations and a standard scheme applying Toeplitz hashing. These improvements enable our MTJ-based TRNGs to ensure security at both the device and application levels. Integrating the TRNGs into GAI greatly improves security, particularly against black-box attacks. By combining our TRNGs with complementary security strategies, such as robust access controls, we believe it can benefit GAI security more broadly. Furthermore, our lightweight RNG scheme is well-suited for IoT applications, offering high-quality true randomness with minimal energy overhead. This enhancement scheme could also be extended to other TRNG technologies.

## 4. Experimental Section/Methods

*Sample Growth and Device Fabrication*: Thin-film samples with the structure of substrate/[W (3)/Ru (10)]$_2$/W (3)/Pt (3)/Co (0.25)/Pt (0.2)/[Co (0.25)/Pt (0.5)]$_5$/Co (0.6)/Ru (0.85)/Co (0.6)/Pt (0.2)/Co (0.3)/Pt (0.2)/Co (0.5)/W (0.3)/CoFeB (0.9)/MgO (0.85)/CoFeB (1.2)/W (0.4)/CoFeB (0.8)/MgO (0.7)/Ta (3)/Ru (7)/Ta (5) are deposited on Si substrates with a 300 nm thermal oxide layer via direct current (DC) magnetron sputtering for metallic layers and radio frequency (RF) magnetron sputtering for the MgO layers. Deposition is conducted at room temperature with a base pressure below $2\times10^{-8}$ Torr. All thickness values given in parentheses are in nanometers. To fabricate the MTJs, bottom electrode structures 10 μm wide are initially patterned using photolithography and Ar ion milling. MTJ pillars with diameters of ~80 nm are patterned via electron-beam lithography. After ion milling, an encapsulation layer of Si$_3$N$_4$ is deposited in-situ without breaking vacuum using RF magnetron sputtering. Finally, top electrode structures measuring 10 μm in width are patterned using photolithography, and the top electrode layers composed of Ta (5 nm)/Cu (40 nm) are deposited via DC magnetron sputtering.

*Printed Circuit Board Design*: A 16-channel DAC (AD5767) is used to generate bipolar analog input voltages ($V_{dd}$) for the 16 MTJs, providing individual control for each MTJ. A 16-channel ADC (MAX11131), along with fixed-value sense resistors, samples the outputs from each MTJ. The sampled raw data are then sent to an FPGA (VCU118) for storage. The polarity of the $V_{dd}$ voltage determines the current direction through each MTJ, enabling reset and perturb operations. By adjusting the amplitude of $V_{dd}$, the current amplitude flowing through each MTJ can also be tuned. Communication between the DAC, ADC, and FPGA is facilitated via the serial peripheral interface (SPI) protocol. The FPGA is programmed and controlled using Verilog. During our experiments, the system operates at a main frequency of 100 kHz. Given the parallelism of 16 MTJs, the system achieves a sampling rate of 1.6 million random bits per second.

*MTJ-based System Characterization*: To generate random bits using STT-MTJ computing cells, we first apply a fixed negative $V_{dd}$ voltage (reset pulse) to deterministically set each MTJ to the AP state. Subsequently, we apply a positive $V_{dd}$ voltage (perturb pulse) to induce stochastic switching from the AP state to the P state. The switching probability is controlled by adjusting the amplitude of the positive perturb pulse. In our experiments, we aim for a 50% switching probability; hence, the perturb pulse amplitude is fine-tuned and fixed at a specific value for each MTJ. The ADC then samples the output voltage ($V_{out}$) from each MTJ, and the sampled values are sent to the FPGA. The FPGA compares each sampled voltage with a pre-characterized threshold voltage ($V_{th}$). If $V_{out}$ exceeds $V_{th}$, the bit is stored as '1'; otherwise, it is stored as '0'. Due to the non-volatile nature of MTJ devices, a reset pulse is required before generating each random bit. By simultaneously controlling 16 MTJs in parallel using the FPGA, we acquire 16 bits at one cycle (reset-perturb). In total, a raw random bit sequence consisting of $10^8$ bits is recorded.

**Figures and Tables**

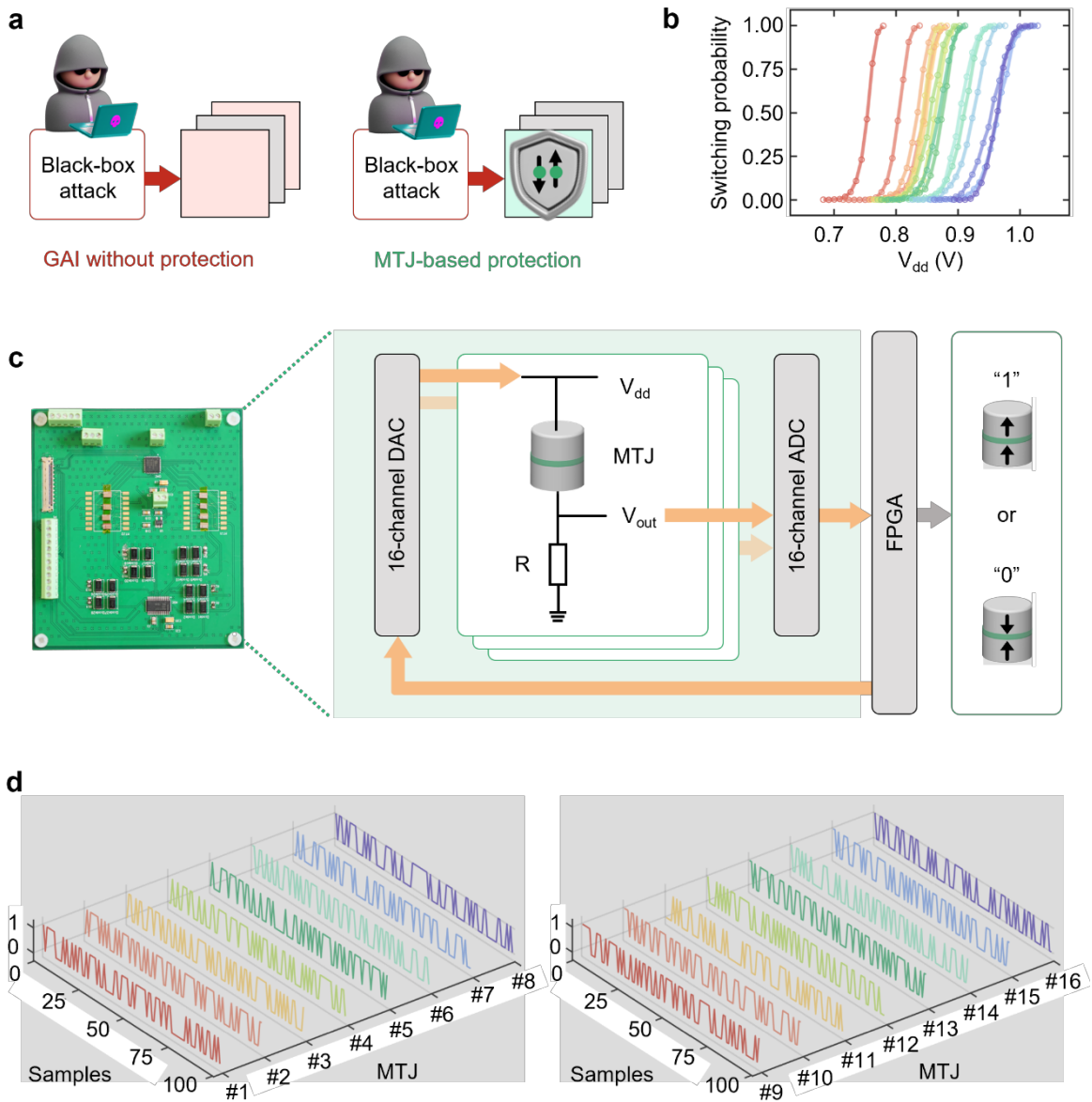

**Figure 1. Experimental setup for the MTJ-based computing system.** a) Motivation of this study showing the scenario without protection (left panel) and MTJ-based true randomness as a defense against GAI security risks (right panel). b) Switching probability versus the input voltage $V_{dd}$. Different colors for the data lines corresponding to different MTJs. c) Photograph and diagram of the MTJ-based computing system. The system consists of 16 computing MTJ cells, a 16-channel digital-to-analog converter (DAC), and a 16-channel analog-to-digital converter (ADC) for writing to and reading from each MTJ cell. The components are controlled by an FPGA via a serial peripheral interface (SPI). d) Output of MTJs #1 to #16 when each cell is set to a 50% switching probability.

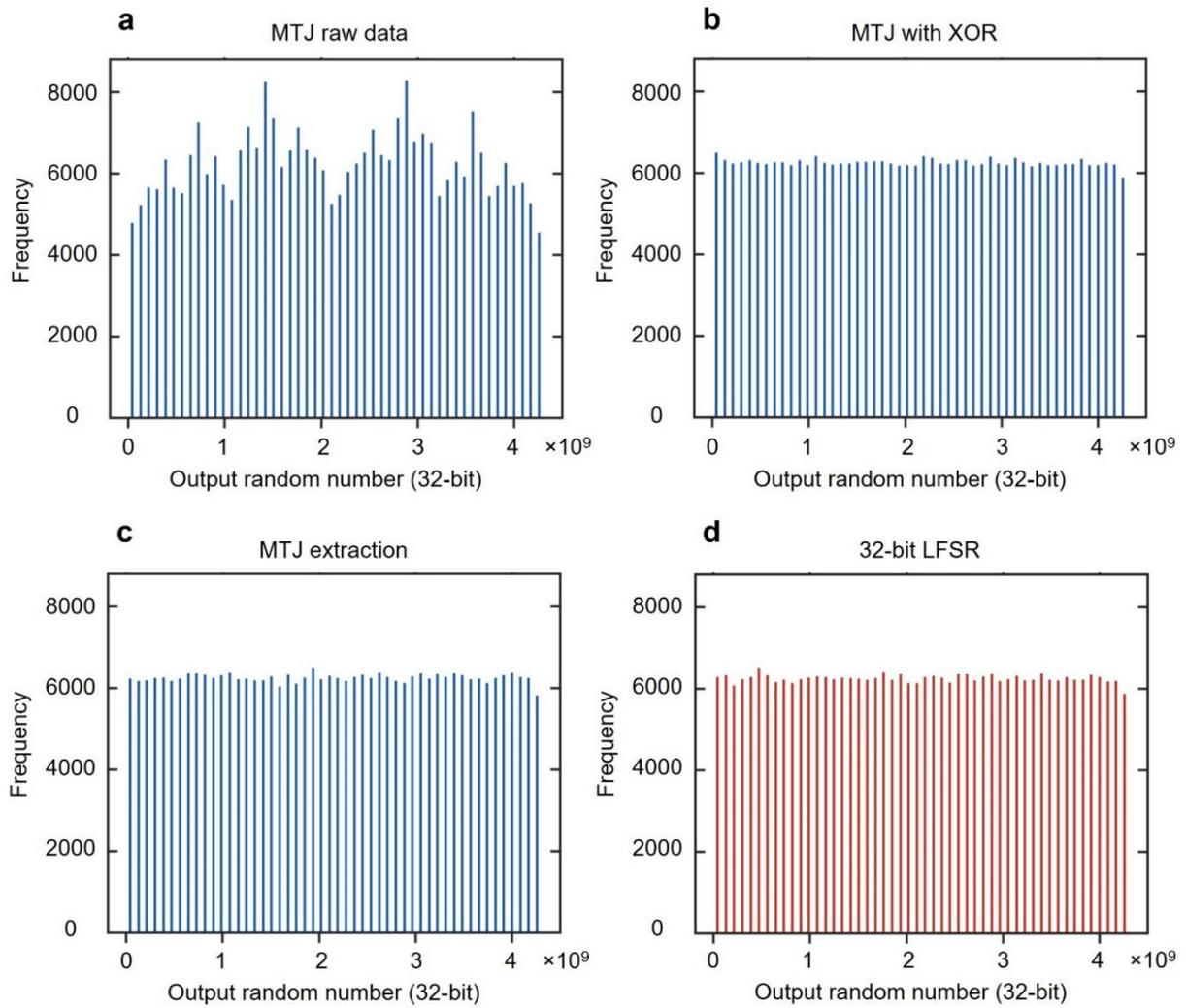

**Figure 2. Verifying the functionality of four RNGs.** a) MTJ raw data. b) MTJ with XOR. c) MTJ extraction using Toeplitz hashing. d) 32-bit linear feedback shift register (LFSR).

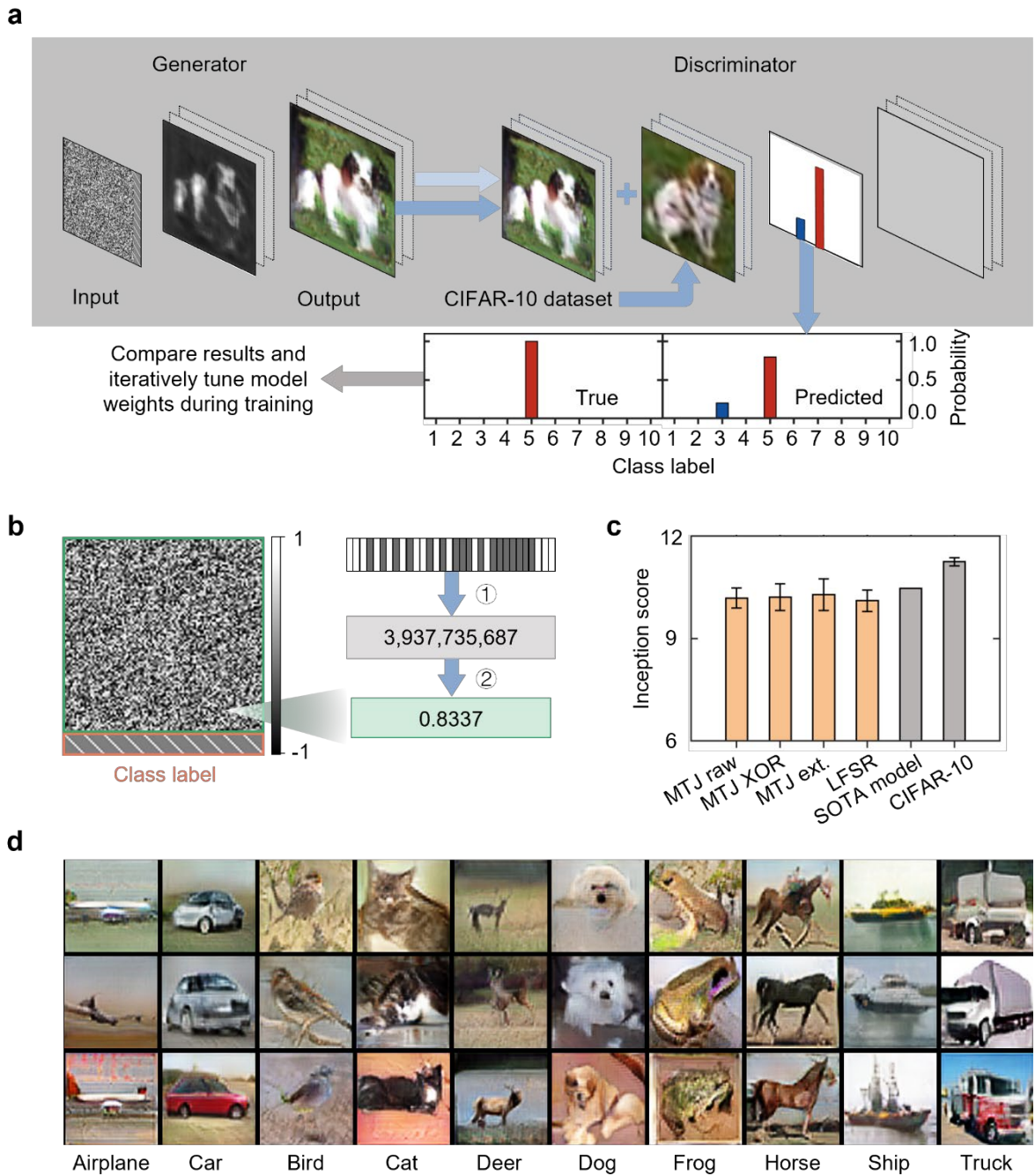

**Figure 3. Image generation with MTJ-based TRNGs.** a) The basic structure of a GAN includes a generator, which expands random input (latent code) through intermediate layers to generate the final output, and a discriminator, which takes data from both the training dataset and the generator to perform classification tasks. The predicted results are compared with ground truth to adjust the model weights during training. Class labels 1-10 correspond to the ten CIFAR-10 categories: airplane, car, bird, cat, deer, dog, frog, horse, ship, and truck. b) An example of the random input to the GAN and how it is produced using hardware RNGs. In the upper right corner, black represents binary bit 0 and white represents 1. c) Inception score (IS) of the image generation results. d) Images generated using the MTJ with XOR RNG (from left to right, class labels 1-10).

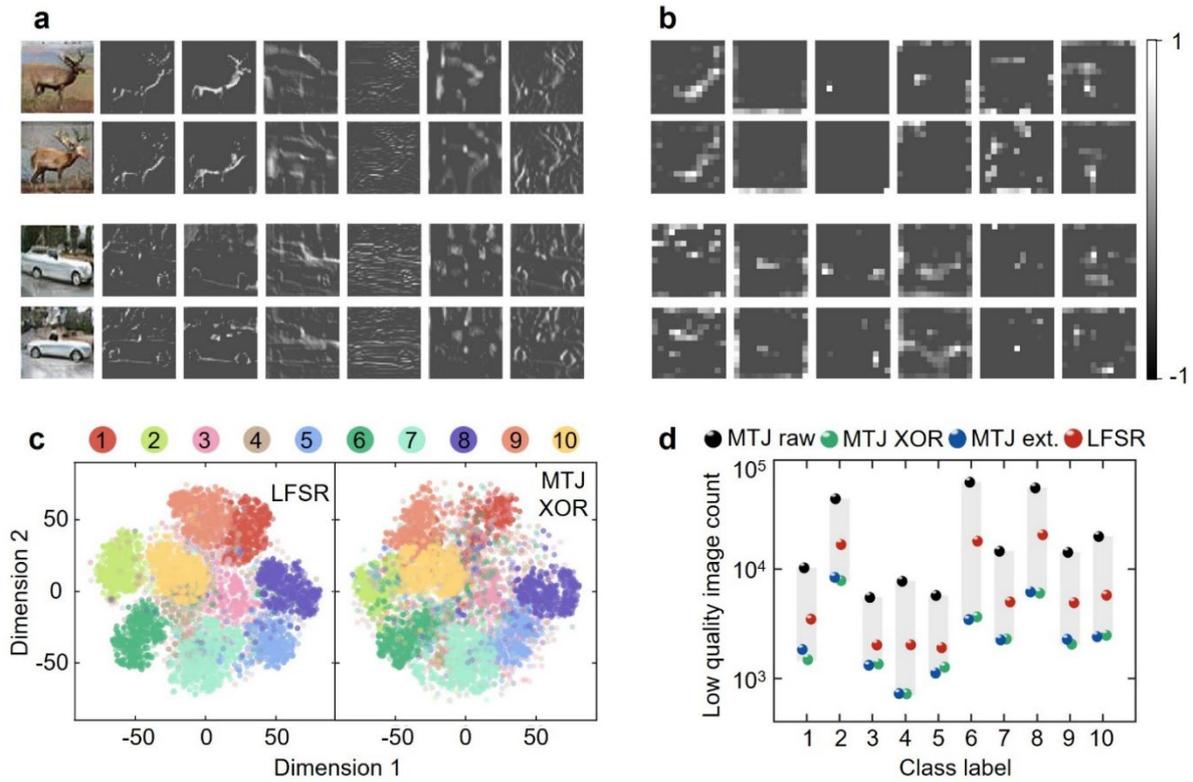

**Figure 4. Intra-class diversity analysis of RNGs.** a) and b) Visually similar image pairs observed using the LFSR RNG and their corresponding extracted feature maps. a) Original image pairs and low-level features. b) High-level features. The upper pair has a similarity score of 0.199, and the lower pair has a score of 0.149. c) t-distributed stochastic neighbor embedding (t-SNE) analysis of the generated images. The left panel shows plot from the LFSR, and the right panel from the MTJ with XOR. In both plots, each data point represents a generated image, and different colors indicate different classes from 1 to 10. The numbers on the colored dots correspond to the class labels (1-10), representing the ten CIFAR-10 categories. d) The count of low-quality images (similarity score < 0.3) within the same class for all class labels and all RNGs. Scatter points represent the average count for each RNG, while the shaded area indicates the full range of counts observed across different RNGs.

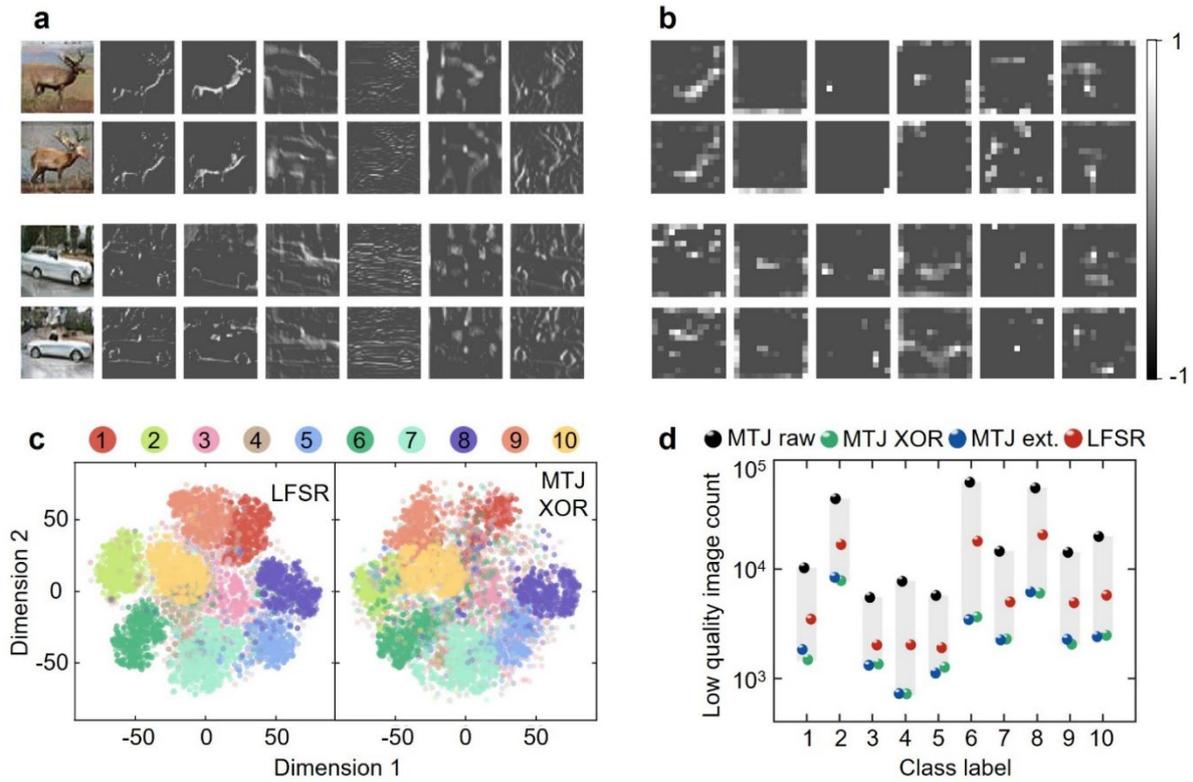

**Figure 4. Intra-class diversity analysis of RNGs.** a) and b) Visually similar image pairs observed using the LFSR RNG and their corresponding extracted feature maps. a) Original image pairs and low-level features. b) High-level features. The upper pair has a similarity score of 0.199, and the lower pair has a score of 0.149. c) t-distributed stochastic neighbor embedding (t-SNE) analysis of the generated images. The left panel shows plot from the LFSR, and the right panel from the MTJ with XOR. In both plots, each data point represents a generated image, and different colors indicate different classes from 1 to 10. The numbers on the colored dots correspond to the class labels (1-10), representing the ten CIFAR-10 categories. d) The count of low-quality images (similarity score < 0.3) within the same class for all class labels and all RNGs. Scatter points represent the average count for each RNG, while the shaded area indicates the full range of counts observed across different RNGs.

**Table 1. NIST statistical test suite results of four RNGs.** The vertical axis lists the tests in the suite. Each cell value indicates the pass ratio. Green indicates a failure, and yellow indicates that all sub-tests passed. Each number in the cells represents the uniformity-of-p-values. For tests consisting of multiple sub-tests, the results are shown as either PASS or FAIL, with FAIL meaning that one or more sub-tests did not pass.

| Test name | MTJ raw data | MTJ with XOR | MTJ extraction | 32-bit LFSR |
|---|---|---|---|---|
| Frequency | 0.00 | 0.90 | 0.60 | 0.22 |
| Block frequency | 0.00 | 0.22 | 0.01 | 0.16 |
| Runs | 0.00 | 0.40 | 0.22 | 0.16 |
| Longest run | 0.00 | 0.55 | 0.18 | 0.40 |
| Rank | 0.51 | 0.51 | 0.55 | 0.00 |
| FFT | 0.00 | 0.55 | 0.25 | 0.44 |
| Non-overlapping template | FAIL | PASS | PASS | PASS |
| Overlapping template | 0.00 | 0.44 | 0.18 | 0.02 |
| Universal | 0.00 | 0.37 | 0.18 | 0.15 |
| Linear complexity | 0.37 | 0.47 | 0.06 | 0.00 |
| Serial | FAIL | PASS | PASS | PASS |
| Approximate entropy | 0.00 | 0.68 | 0.13 | 0.01 |
| Cumulative sum | FAIL | PASS | PASS | PASS |
| Random excursions | FAIL | PASS | PASS | PASS |
| Random excursions variant | FAIL | PASS | PASS | PASS |